%% file: main.tex
\documentclass[10pt,twocolumn,letterpaper]{article}

\usepackage{cvpr}              

\definecolor{cvprblue}{rgb}{0.21,0.49,0.74}
\usepackage[pagebackref,breaklinks,colorlinks,allcolors=cvprblue]{hyperref}
\usepackage{booktabs}

\title{Multi-Object Tracking Retrieval with LLaVA-Video: \\
A Training-Free Solution to MOT25-StAG Challenge}

\author{Yi Yang$^{1}$ \quad Yiming Xu$^{1}$ \quad Timo Kaiser$^{1}$ \quad Hao Cheng$^{2}$ \quad Bodo Rosenhahn$^{1}$ \quad Michael Ying Yang$^{3}$\\
$^{1}$Leibniz University Hannover \quad $^{2}$University of Twente \quad $^{3}$University of Bath\\
{\tt\small \{yangyi,kaiser,rosenhahn\}@tnt.uni-hannover.de, yiming.xu@ikg.uni-hannover.de,} \\
{\tt\small
h.cheng-2@utwente.nl, \
myy35@bath.ac.uk
}}

\begin{document}
\maketitle
\input{sec/0_abstract}    
\input{sec/1_intro}
\input{sec/2_baseline}
\input{sec/3_method}
\input{sec/4_results}
\input{sec/5_discussion}
\input{sec/6_acknowledgement}
{
    \small
    \bibliographystyle{ieeenat_fullname}
    \bibliography{main}
}


\end{document}

%% file: sec/0_abstract.tex
\begin{abstract}
In this report, we present our solution to the MOT25-Spatiotemporal Action Grounding (MOT25-StAG) Challenge. The aim of this challenge is to accurately localize and track multiple objects that match specific and free-form language queries, using video data of complex real-world scenes as input. We model the underlying task as a video retrieval problem and present a two-stage, zero-shot approach, combining the advantages of the SOTA tracking model FastTracker and Multi-modal Large Language Model LLaVA-Video. On the MOT25-StAG test set, our method achieves m-HIoU and HOTA scores of 20.68 and 10.73 respectively, which won second place in the challenge. 
\end{abstract}

%% file: sec/1_intro.tex
\section{Introduction}
\label{sec:intro}

The MOT25-StAG competition \cite{hannan2025svagbenchlargescalebenchmarkmultiinstance} introduces a novel benchmark that extends the traditional multi-object tracking (MOT) task \cite{leal2015motchallenge,dendorfer2020mot20}. Unlike standard MOT tasks, which focus solely on detecting and consistently tracking objects across frames, this challenge integrates temporal action localization with tracking, guided by natural language queries. This unification requires models not only to follow objects in space and time but also to understand and ground them according to complex, free-form action descriptions.

The novel task composition introduces several unique challenges. First, models must bridge vision and language, interpreting ambiguous or diverse textual queries and aligning them with visual evidence. Second, they must jointly handle the temporal localization of when the described action occurs and the spatial tracking of the relevant objects. Both tasks can be highly dynamic in real-world videos. Finally, the integration of multiple datasets with dense manual annotations raises the bar for generalization, demanding robustness to diverse scenes, crowded environments, and fine-grained distinctions between similar actions. 

Our key observation is that MOT25-StAG queries demand comprehensive video-level understanding, often involving relational or sequential reasoning (e.g., “track the dog that is the first to go toward the car”). Therefore, we  propose a two-stage method where in the first stage all observable objects are tracked, and in the second stage every track is captioned by LLaVA-Video. Tracks that matches the target queries are retrieved using cosine similarity between the query and the caption. 


%% file: sec/2_baseline.tex
\section{Related Methods}
\label{sec:baseline}

\begin{figure*}[t]
    \centering
    \includegraphics[width=\textwidth]{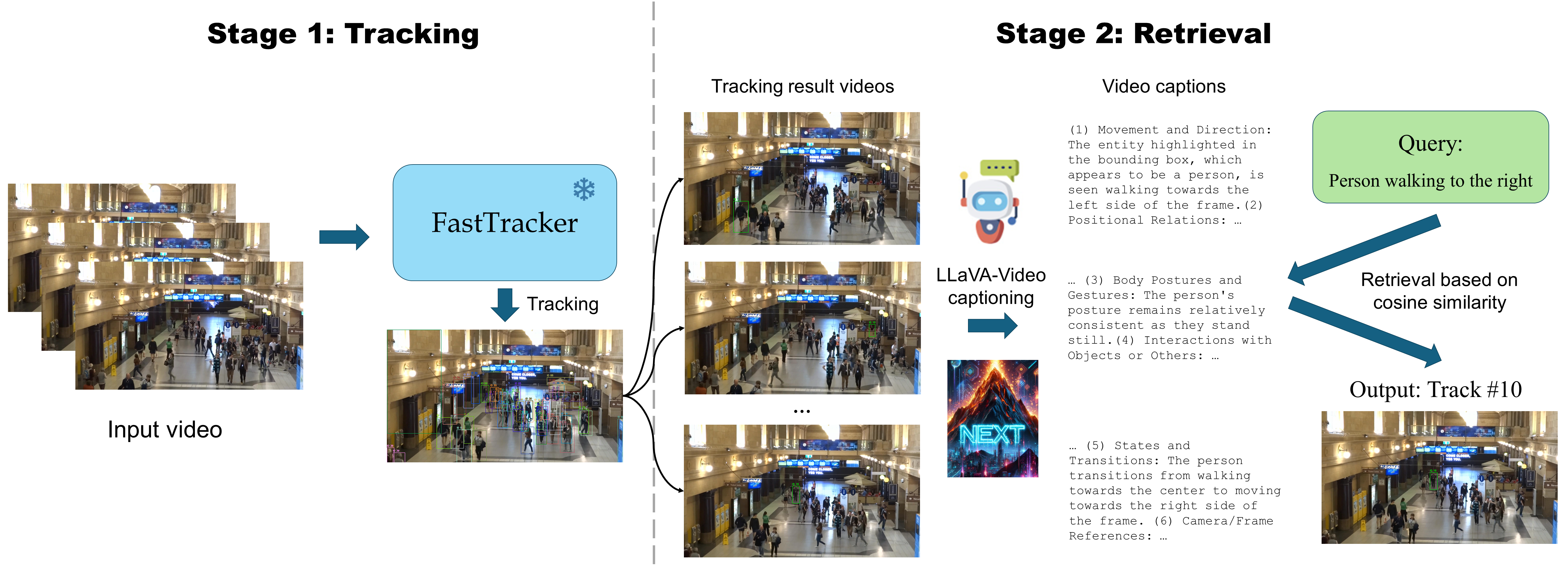}
    \caption{Our two-stage training-free framework for spatiotemporal action grounding. In the first stage, we track all the objects with a pre-trained tracking model FastTracker, and generate tracking results, one video for each track. In the second stage, we caption the resulting tracking result videos with LLaVA-Video. Videos are retrieved based on the similarity of the caption and the query. }
    \label{fig:framework}
\end{figure*}

Recent progress in spatial-temporal video action grounding has been driven by advances in referring multi-object tracking and video temporal grounding. Referring multi-object tracking is the task of detecting and tracking multiple objects in a video according to a natural language description. TempRMOT \cite{zhang2024bootstrapping} addresses this task by proposing a temporally enhanced query-based framework. At each timestamp, TempRMOT takes in visual features from the video frame and linguistic features from RoBERTa \cite{liu2019roberta}. These are projected into the same dimension and passed into a fusion encoder that applies cross-attention, aligning the modalities. TempRMOT further incorporates a memory-based temporal module to strengthen long-term spatial-temporal modeling, achieving state-of-the-art results in associating objects with natural language queries across time. 

FlashVTG \cite{cao2025flashvtg}, on the other hand, introduces a novel framework for text-guided video temporal grounding, the task of localizing the precise temporal segment of a video that matches a natural language description. FlashVTG uses the CLIP \cite{radford2021learning} text encoder together with GloVe embeddings for word-level features, and CLIP image encoder and SlowFast \cite{feichtenhofer2019slowfast} for visual features from video clips. After encoding, video and text features are projected into the same space and then fused using an Adaptive Cross-Attention module. With a further Temporal Feature Layering module to better capture multi-scale temporal information, FlashVTG significantly improves the retrieval of short, fine-grained video moments. 

Although both TempRMOT and FlashVTG represent strong baselines in their respective domains, they do not directly align with the requirements of the MOT25-StAG challenge. TempRMOT is well-suited for spatial tracking but begins from frame-level object detection and lacks global temporal reasoning, whereas FlashVTG is strong in temporal grounding but does not support spatial localization or object-level tracking. Therefore, we propose a two-stage tracking and retrieval pipeline as introduced in the following section. 

%% file: sec/3_method.tex
\section{Method}
\label{sec:method}

Figure 1 illustrates the general workflow of our method. In stage-1, we aim to obtain tracks for all objects seen in the video employing the off-the-shelf SOTA tracking model \textit{FastTracker} \cite{hashempoor2025fasttracker}. 
We predict multiple potential object instance trajectories for an input video. Based on the trajectory estimations, we generate a set of output videos in which only one instance is highlighted by surrounding green bounding boxes, respectively.
Then, in stage-2, we utilize LLaVA-Video \cite{zhang2025llava} to generate captions for every tracking result video. For each query, we retrieve the most relevant captions and thus the tracking results. 
Our pipeline only includes substitutable pretrained models and is not fine-tuned, which we hope to provide an adaptable baseline for future research. 
We also acknowledge that unfortunately due to limited time we were only able to apply our method to the MOT17 \cite{leal2015motchallenge} and MOT20 \cite{dendorfer2020mot20} datasets. For OVIS \cite{qi2022occluded} dataset, we simply used the results from TempRMOT. 

\subsection{Tracking Stage: Implementation Details}

Our proposed retrieval pipeline relies on strong trajectory proposals, which are later used to generate language descriptions. Thus, we employ the current state-of-the-art tracking method \textit{FastTracker}~\cite{hashempoor2025fasttracker} to predict high-quality tracking proposals. 
\textit{FastTracker} is built on the lightweight tracking-by-detection framework \textit{ByteTrack}~\cite{zhang2022bytetrack}, which is extended with a re-identification module to handle occlusions and an environment model that allows refining motion predictions of the used Kalman filter. 
We use the hyperparameters shown in Table~\ref{tab:parameter} and the official weights released for the MOT17 and MOT20 datasets..


\begin{table}[h]
\centering
\resizebox{\linewidth}{!}{%
\begin{tabular}{ccc}
\toprule
Name & Value & Description\\
\midrule

\textit{track\_thresh} & 0.7 & Minimum detection score to initialize/update track. \\
\textit{track\_buffer} & 30 & Frames a tracklet survives without detection.\\
\textit{match\_thresh} & 0.85 & IOU threshold for associating detections to tracks.\\
\textit{min\_box\_area} & 100 & Minimum box area considered for tracking.\\
\textit{reset\_velocity\_offset\_occ} & 5 & Velocity smoothing offset when occluded.\\
\textit{reset\_pos\_offset\_occ} & 3 & Position smoothing offset for occluded tracks.\\
\textit{enlarge\_bbox\_occ} & 1.1 & Enlargement for occluded bounding boxes.\\
\textit{dampen\_motion\_occ} & 0.89 & Dampening factor for velocity of occluded tracks.\\
\textit{active\_occ\_to\_lost\_thresh} & 10 & Max frames occluded before marked lost.\\
\textit{init\_iou\_suppress} & 0.8 & IOU suppression to avoid duplicate track init.\\    
\bottomrule
\end{tabular}
}
\caption{Parameterization of \textit{FastTracker} used in our tracking pipeline.}
\label{tab:parameter}
\end{table}

\begin{table*}[!htbp]
\centering
\resizebox{\linewidth}{!}{%
\begin{tabular}{c|ccccccccccccccccccc}
\hline
Method & m-HIoU & HOTA & mIoU & DetA & AssA & DetRe & DetPr & AssRe & AssPr & LocA & R1@0.1 & R1@0.3 & R1@0.5 & R5@0.1 & R5@0.3 & R5@0.5 & R10@0.1 & R10@0.3 & R10@0.5 \\
\hline

SVAGFormer & 14.15 & 9.16 & 19.14 & 4.09 & 27.70 & 7.12 & 7.58 & 38.45 & 41.45 & 73.02 & 38.59 & 24.57 & 17.75 & 64.79 & 40.57 & 26.23 & 71.84 & 44.54 & 30.43 \\

\textbf{Ours} & 20.68 & 10.73 & 30.63 & 4.07 & 41.69 & 7.86 & 7.29 & 71.33 & 51.30 & 84.03 & 71.14 & 33.65 & 22.89 & 71.14 & 33.65 & 22.89 & 71.14 & 33.65 & 22.89 \\
\hline
\end{tabular}%
}
\caption{Evaluation results on MOT25-StAG test set. Larger values are better. }
\end{table*}

\subsection{Retrieval Stage: Implementation Details}

We use the LLaVA-Video-7B-Qwen2 checkpoint. For each tracking result video, we evenly extract 24 frames for captioning. We instruct LLaVA-Video to focus on the action of the tracked entity. To ensure that the generated captions align better with the language queries in MOT25-StAG, we first use ChatGPT (GPT-5) to summarize the aspects of the actions described by the queries. ChatGPT returns the following 6 aspects as shown in our prompt for LLaVA-Video:

\texttt{Please describe the activities of the entity highlighted in the bounding box, including the following aspects: (1) Movement and Direction; (2) Positional Relations; (3) Body Postures and Gestures; (4) Interactions with Objects or Others; (5) States and Transitions; (6) Camera/Frame References. }

For retrieval, we utilize the Large Language Model all-MiniLM-L6-v2 \cite{reimers-2019-sentence-bert} to embed all language queries and video captions. The embedding dimension is 384. In the embedding space, for each query, we retrieve the most relevant top-k=10 captions using cosine similarity. 

%% file: sec/4_results.tex
\section{Results}
\label{sec:results}

\begin{figure}[h]
    \centering
    \includegraphics[width=0.45\textwidth]{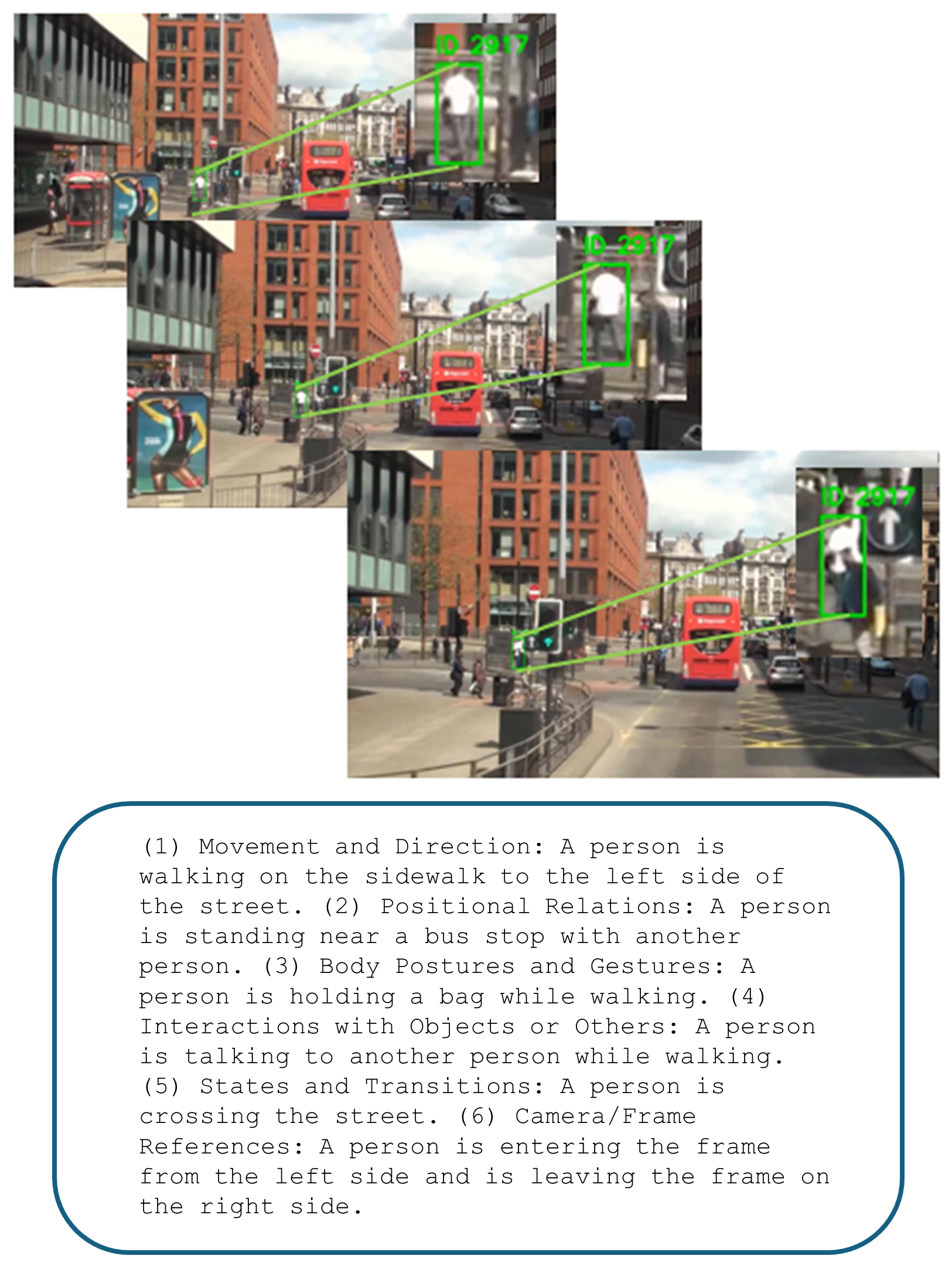}
    \caption{A sample video caption generated by LLaVA-Video, instructed to focus on the action of the person tracked and highlighted in a bounding box. }
    \label{fig:caption}
\end{figure}

We evaluated our method on the MOT25-StAG test set. 
Evaluation metrics include Higher Order Tracking Accuracy (HOTA) \cite{luiten2021hota},  mean Intersection over Union (mIoU), detection accuracy (DetA), association accuracy (AssA), and top-k moment retrieval with confidence threshold X (R-k@X). The main metric to rank submissions in the competition is m-HIoU, which is the average of HOTA and mIoU. Larger values are better.

TempRMOT returns 1322 tracks on the OVIS dataset. On the other hand, FastTracker detects 44, 204, 190, 126, 131, 167, 233, 1199, and 1339 tracks for videos MOT17-01, MOT17-03, MOT17-06, MOT17-07, MOT17-08, MOT17-12, MOT17-14, MOT20-03, and MOT20-05, respectively. 

Figure 2 shows a sample of a tracking result video and its corresponding caption generated by LLaVA-Video. 

Since our retrieval returns the 10 most relevant tracks per query, we obtain 4990 and 4320 tracks for MOT17 and MOT20, respectively. Together with the 1322 tracks for OVIS, our method yields an m-HIoU of 20.68 and a HOTA of 10.73 on the MOT25-StAG test set. Full evaluation results are shown in Table 2.  

%% file: sec/5_discussion.tex
\section{Discussion}
\label{sec:discussion}

Our method significantly outperforms the official baseline SVAGFormer \cite{hannan2025svagbenchlargescalebenchmarkmultiinstance}, as shown in Table 2. In particular, our method performs notably well in AssA. This is because in our two-stage setting, temporal grounding does not interfere with tracking results, avoiding potential error propagation from inaccurate grounding to tracking. We also achieve better results in terms of R1@X, demonstrating the capability of LLaVA-Video to understand the scene and generate precise captions for action. Altogether, these advantages enable our method to have better overall performance and obtain a higher m-HIoU than the official baseline. 

We also showcase a failure case in Figure 3. A human observer would say ``the highlighted person is moving towards the camera". However, LLaVA-Video says ``person is moving to the center of the crowd", which can lead to misalignment in the retrieval stage. In addition, we also observe model hallucination, which says "the person is captured in a single frame". A potential solution is to fine-tune LLaVA-Video so that the model's knowledge is better aligned with the semantic space of target queries. 

\begin{figure}[h]
    \centering
    \includegraphics[width=0.45\textwidth]{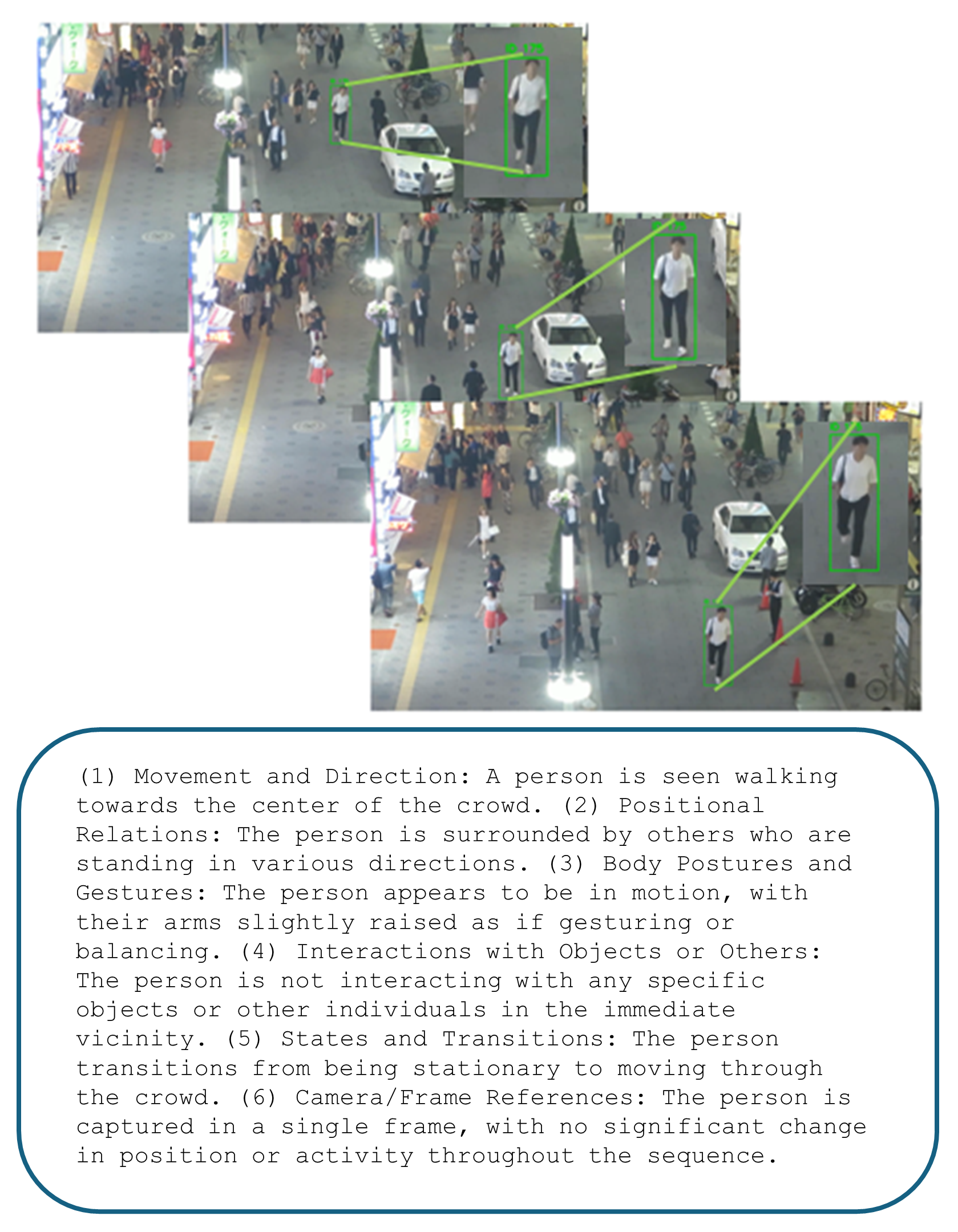}
    \caption{A failure case of video captioning. The person's precise action of moving towards the camera is not captured, and there is a hallucination that "the person is captured in a single frame". }
    \label{fig:failure}
\end{figure}

We acknowledge that there has been concurrent work on referring multi-object tracking for autonomous driving scenarios \cite{chamiti2025refergpt}. This shows growing interest in the community on such flexible training-free methods. 

To summarize, in this report, we present a zero-shot approach for the MOT25-StAG challenge. Our two-stage design utilizes and combines the advantages of tracking model \textit{FastTracker} and language model LLaVA-Video, which provides a flexible baseline for the spatiotemporal video action grounding task. We hope that our method can inspire future research in this area. 

%% file: sec/6_acknowledgement.tex
\section{Acknowledgment}
\label{sec:acknowledgment}

We thank Patrick Glandorf for presenting our work at ICCV'25 Workshop on Benchmarking Multi-Target Tracking and his valuable suggestions. 